%% file: paper.tex
\documentclass[a4paper,UKenglish,cleveref, autoref, thm-restate]{lipics-v2021}
\pdfoutput=1

\input{macros}

\title{Cleaning Inconsistent Data in Temporal \textit{DL-Lite} Under Best Repair Semantics}

\author{Sabiha Tahrat}{LIPADE, Universit\'e de Paris, France}{sa.tahrat@gmail.fr}{}{}

\author{Salima Benbernou}{LIPADE, Universit\'e de Paris, France}{salima.benbernou@u-paris.fr}{}{}

\author{Mourad Ouziri}{LIPADE, Universit\'e de Paris, France}{mourad.ouziri@u-paris.fr}{}{}

\authorrunning{Tahrat et. al}
\Copyright{Sabiha Tahrat and Salima Benbernou and Mourad Ouziri} 



\category{} 

\relatedversion{}

\EventEditors{Carlo Combi, Johann Eder , Mark Reynolds}
\EventNoEds{2}
\EventLongTitle{The 28th International Symposium on Temporal Representation and Reasoning (TIME 2021)}
\EventShortTitle{TIME 2021}
\EventAcronym{TIME 2021}
\EventYear{2021}
\EventDate{September 27--29, 2021}
\EventLocation{University of Klagenfurt, Austria}
\EventLogo{}
\SeriesVolume{}
\ArticleNo{}

\begin{document}
\nolinenumbers
\maketitle  

\input{0-abstract.tex}

\input{1-introduction.tex}

\input{2-relatedworks.tex}
\input{3-preliminaries}

\input{4-translation.tex}

\input{5-detection.tex}

\input{7-conclusion.tex}
\newcommand{\SortNoop}[1]{}
\bibliographystyle{plainurl}
\bibliography{RefPaper}

\end{document}

%% file: macros.tex
\usepackage{xspace}
\usepackage{tikz}
\usepackage{wasysym}
\usepackage{amsmath,amsthm,mathtools,amssymb,amsfonts}
\newcommand{\nb}[1]{}

\newcommand{\eset}{\emptyset}

\newcommand{\sqs}{\sqsubseteq}

\newcommand{\Next}{\ensuremath{\fullmoon}}

\newcommand{\D}{\Diamond}

\newcommand{\B}{\Box}

\newcommand{\monodic}{\ensuremath{\mathop{\ooalign{$\Box$ \cr \kern0.55ex \raisebox{0.2ex}{\scalebox{0.5}{$1$}}}\rule{0pt}{1.5ex} \kern-0.7ex}}\xspace}

\newcommand{\DL}{\ensuremath{\text{DL}}\xspace}

\newcommand{\DLite}{\textsl{DL-Lite}\xspace}

\newcommand{\DLiteA}{\ensuremath{\DLite_\textit{A}}\xspace}

\newcommand{\DLitebnhn}{\ensuremath{\DLite_\textit{bool}^{\cal (NH)}}\xspace}

\newcommand{\NC}{\ensuremath{{\sf N_C}}\xspace}
\newcommand{\NI}{\ensuremath{{\sf N_I}}\xspace}
\newcommand{\NR}{\ensuremath{{\sf N_R}}\xspace}

\newcommand{\NGl}{\ensuremath{\textsf{N}_{\textsf{G}}\xspace}}
\newcommand{\NL}{\ensuremath{\textsf{N}_{\textsf{L}}\xspace}}

\newcommand{\cl}{\ensuremath{\mathsf{cl}}\xspace}

\newcommand{\TDL}{\ensuremath{\text{TDL}}\xspace}

\newcommand{\TDLLite}{\ensuremath{\textsl{TDL-Lite}}\xspace}
\newcommand{\TDLLiteN}{\ensuremath{\textsl{T}^{\Nbb}\textsl{DL-Lite}}\xspace}

\newcommand{\Rdiamond}{\Diamond_{\!\scriptscriptstyle F}}
\newcommand{\Rbox}{\Box_{\!\scriptscriptstyle F}}

\newcommand{\SVdiamond}{\mathop{\ooalign{$\Diamond$ \cr \kern0.5ex
    \raisebox{0.35ex}{\scalebox{0.7}{$*$}}} \kern-0.9ex}}
\newcommand{\SVbox}{\mathop{\ooalign{$\Box$ \cr \kern0.42ex
    \raisebox{0.35ex}{\scalebox{0.7}{$*$}}}\rule{0pt}{1.5ex} \kern-0.7ex}}

\newcommand{\NP}{\textsc{NP}}

\newcommand{\Mmf}{\ensuremath{\mathfrak{M}}\xspace}

\newcommand{\Wmf}{\ensuremath{\mathfrak{W}}\xspace}

\newcommand{\Amc}{\ensuremath{\mathcal{A}}\xspace}
\newcommand{\Bmc}{\ensuremath{\mathcal{B}}\xspace}
\newcommand{\Cmc}{\ensuremath{\mathcal{C}}\xspace}
\newcommand{\Dmc}{\ensuremath{\mathcal{D}}\xspace}

\newcommand{\Imc}{\ensuremath{\mathcal{I}}\xspace}

\newcommand{\Kmc}{\ensuremath{\mathcal{K}}\xspace}
\newcommand{\Lmc}{\ensuremath{\mathcal{L}}\xspace}

\newcommand{\Rmc}{\ensuremath{\mathcal{R}}\xspace}

\newcommand{\Tmc}{\ensuremath{\mathcal{T}}\xspace}

\newcommand{\Xmc}{\ensuremath{\mathcal{X}}\xspace}

\newcommand{\N}{\mathbb{N}} 
\newcommand{\Nbb}{\ensuremath{\mathbb{N}}\xspace}

\newcommand{\A}{\ensuremath{\mathcal{A}}}
\newcommand{\K}{\ensuremath{\mathcal{K}}}
\newcommand{\I}{\ensuremath{\mathcal{I}}}
\newcommand{\T}{\ensuremath{\mathcal{T}}}
\newcommand{\Shoin}{\ensuremath{\mathcal{SHOIN(D)}}\xspace}

\newcommand{\tdllitefpx}{$T_{FPX}DL$-$Lite^{\mathcal{N}}_{bool}$}



%% file: 0-abstract.tex
\begin{abstract} 
\noindent

In this paper, we address the problem of handling inconsistent data in Temporal Description Logic (TDL) knowledge bases. Considering the data part of the knowledge base as the source of inconsistency over time, we propose an ABox repair approach. This is the first work handling the repair in TDL Knowledge bases. To do so, our goal is twofold: 1) detect temporal inconsistencies and 2) propose a data temporal reparation. For the inconsistency detection, we propose a reduction approach from TDL to DL which allows to provide a tight NP-complete upper bound for TDL concept satisfiability and to use highly optimised DL reasoners that can bring precise explanation (the set of inconsistent data assertions). Thereafter, from the obtained explanation, we propose a method for automatically computing the best repair in the temporal setting based on the allowed rigid predicates and the time order of assertions.
\end{abstract}


%% file: 1-introduction.tex
\section{Introduction}
\label{sec:intro}





The Ontology Web Language (OWL) is the ontology language recommended by the W3C that can be used to model knowledge domains. OWL is derived from the well known Description Logics (DLs) \cite{handbookDL} which provide the basic representation features of OWL. Despite the expressiveness power of OWL, it cannot fully express the temporal knowledge needed in many applications. Beyond allowing data values to be typed as basic XML Schema dates, times or durations \footnote{https://www.w3.org/TR/xmlschema11-1/}, OWL has a very limited support for temporal information modeling and reasoning. 
\textit{}
Time is crucial because events occur at specific points in time and also the relationships among objects exist over time. The ability to model this temporal dimension is therefore crucial in real- world applications such as banking, medical records and geographical information systems. Thus, the temporalization of DL (TDL) has been studied in \cite{ArtFra2, LutEtAl} but for reasoning tasks such as satisfiability, it is known to be hard. Temporal extensions of the lightweight DL <<DL-Lite>> \cite{DBLP:journals/tocl/ArtaleKRZ14,CalEtAl1} are therefore considered for their low complexity for many reasoning problems. As \TDLLite language we here consider the \TDLLiteN fragment \cite{DBLP:conf/dlog/TahratBAGO20} allowing for: full Boolean connectives, LTL operators \cite{DBLP:conf/focs/Pnueli77} interpreted over \Nbb in the construction of concepts, inverse roles, distinction between local and global roles (aka. rigid roles). And we restrict ourselves to using the two future operators ($\D_F$: eventually in the future and $\B_F$: always in the future) and to specifying functionality on roles or their inverses. Note that, \TDLLiteN is the simpler logic of \tdllitefpx fragment, the most expressive combination of the tractable $\DLite$ family logics with LTL and which is known to be \NP-\textit{complete} \cite{DBLP:journals/tocl/ArtaleKRZ14}.

\TDLLite\ ontologies, also called \emph{temporal knowledge bases} (TKBs), 
are expressed as a finite set of \emph{general concept inclusions GCIs} \emph{TBox}, which is expressed with temporalized concepts, paired with a timestamped factual knowledge \emph{ABox} that represents data at different time points. Therefore, one of the most important challenges in TDL is to deal with inconsistent ontologies where the \emph{ABox} is inconsistent with a satisfiable TBox: a subset of the assertions in the \emph{ABox} contradicts one or more \emph{TBox} assertions.

Then, the ABox is not reliable and must be repaired. The problem of handling inconsistent data in TKBs has not been fully addressed and only focused so far on satisfiability checking  \cite{DBLP:conf/dlog/TahratBAGO20}. The reparation of the inconsistent data assertions in the ABox has not however been addressed yet. 
To the best of our knowledge, we propose the first approach to automatically repair \emph{ABox} over TDL KBs based on the maximal repair semantic. The obtained repair is a maximal subset of the \emph{ABox} that is consistent with the \emph{TBox}. More precisely, we make the following contributions: 

\begin{itemize}
  \item  We present a linear equisatisfiable translation of \TDLLite knowledge bases KBs into DL KBs which allows to provide a tight NP-\textit{complete} upper bound for \TDLLite concept satisfiability and to use highly optimized DL reasoners that can bring precise inconsistency explanations (the set of inconsistent data assertions). 
  \item We adapt the existing automatic repairing approach of an inconsistent DL ontology, based on the maximal repair semantic \cite{DBLP:conf/bigdataconf/BenbernouO17}, for temporal \emph{ABoxes} and we show that this semantic is preserved.
  \item We extend the maximal repair semantic in DL to the temporal setting by associating the detected inconsistent assertions with a temporal weight. This weighting is based on the defined rigid predicates in the \emph {TBox} and the timestamp order of the \emph{ABox} assertions. The repair is computed by removing inconsistent assertions with the lowest weight. 
\end{itemize}

In the following, we present our running example to provide an intuitive overview of the inconsistency detection and the repair over \TDLLite KBs.
\begin{example} \label{ex:1} 
Let's consider the following KB $\K=(\T,\A)$ where $\T$ is a \TDLLite TBox stating that minors and adults are persons, but they are disjoint. Moreover, the concepts person, adult and the role \textit{hasMother} are rigid. The ABox $\A=\A_{1} \cup \A_{2}$ reports $John's$ status in $\A_{1}$ and his mother in $\A_{2}$, at different timestamps:
{\small
\[
\begin{array}{llll}
\T=&\{{\sf Adult} \sqsubseteq {\sf Person},~{\sf Minor} \sqsubseteq
  {\sf Person},~{\sf Person} \sqsubseteq \B_F {\sf Person},~{\sf Adult} \sqsubseteq \B_F {\sf Adult}\\
  & ~~{\sf Minor} \sqcap {\sf Adult} \sqsubseteq \bot,~{\sf Person} \sqsubseteq \exists{\sf hasMother},~~{\sf Funct(hasMother)} \}\\
  \A_{1}=&\{{\sf Person}({\sf John},0), {\sf Minor}({\sf John},1),{\sf Adult}({\sf John},2),{\sf Minor}({\sf John},3),{\sf Minor}({\sf John},4)\}\\
  \A_{2}=&\{{\sf hasMother}({\sf John},{\sf Ana},0),{\sf hasMother}({\sf John},{\sf Eva},1), {\sf hasMother}({\sf John},{\sf Maria},2)\}
\end{array}
\]}
Both $\A_{1}$ and $\A_{2}$ are inconsistent w.r.t \T. In $\A_{1}$, the assertions: {\small${\sf Adult}({\sf John},2)$}, {\small${\sf Minor}({\sf John},3)$} and {\small${\sf Minor}({\sf John},4)$} with the rigidity of adult violate the disjointness between adult and minor. Similarly in $\A_{2}$, John has multiple mothers at different time points which violates the properties (functional and global) of the role {\sf hasMother}.
Then $\A_{1}$ has a maximal repair {\small$\A_{1}'=\{{\sf Person}({\sf John},0),{\sf Minor}({\sf John},1),{\sf Minor}({\sf John},3),{\sf Minor}({\sf John},4)\} $} and $\A_{2}$ has three possible maximal repairs as follows: 
{\small
\[
\begin{array}{lll}
\A_{2}'=\{{\sf hasMother}({\sf John},{\sf Ana},0)\}& \A_{2}''=\{{\sf hasMother}({\sf John},{\sf Eva},1)\}\\ 
\A_{2}'''=\{{\sf hasMother}({\sf John},{\sf Maria},2)\}
\end{array}
\]
}
\end{example}

This paper is structured as follows. In the next section, we sketch works that have been conducted in the context of inconsistent data reparation in knowledge base fields. In section 3, we introduce the syntax and semantics of the temporal DL \TDLLite that formalise our running example. In section 4, we propose a translation to reduce TDL KBs to DL KBs. Based on the obtained DL KBs, in section 5, we perform inconsistency detection and compute the best temporal data repair. Section 6 concludes this paper and presents some future works.



%% file: 2-relatedworks.tex
\section{Related Work} 
The problem of inconsistencies appearing in KBs can be tackled either by repairing the KB, which leads to a consistent version of it \cite{DBLP:journals/tkde/GrecoGZ03}, or by providing the ability to query inconsistent data and get consistent answers (Consistent Query Answering - CQA) \cite{DBLP:conf/pods/ArenasBC99}. These two approaches were applied initially in the context of relational databases and, later, in the context of KBs as well. The repair approaches share the same principle of performing a minimal set of actions (insertions, deletions, updates) over the KB, in order to render it valid with respect to a given set of integrity constraints. They differ in the type of integrity constraints and in the applied actions. However, CQA approaches take into account, at query execution time, all the possible repairs without materializing a repair of the inconsistent KB. They differ in the repair semantics, the type of integrity constraints and ways of computing the repairs. In the following, we will refer to the most notable related works, dealing with the problems of CQA or repairing, that have been proposed for atemporal and temporal KBs.
\subsection{Atemporal KBs Repair} 
The most well-known, and arguably the most natural approach for Consistent Query Answering is the ABox Repair (AR) semantics \cite{DBLP:conf/rr/LemboLRRS10}. It consists of finding all the maximal subsets of the ABox that are consistent with the TBox and thus showing that inconsistency-tolerant instance checking is already intractable. For this reason, the Intersection ABox Repair (IAR) semantics was proposed which is the intersection of all AR-repairs, and is polynomially tractable. Few implemented systems \cite{DBLP:conf/aaai/BienvenuBG14} designing practically efficient consistent query answering systems, that could scale up to billions of data, is still largely open.  
There has been few implementations regarding ABox inconsistency checking. The reasoner QuOnto \cite{DBLP:conf/aaai/AcciarriCGLLPR05} allows to only check the satisfiability of \DLiteA KBs the simplest fragment of \DLite which is at the bases of OWL 2 QL \footnote{https://www.w3.org/TR/owl2-profiles/}. 
A preliminary ABox cleaner system QuAC implemented within QuOnto is reported by \cite{DBLP:conf/dlog/MasottiRR11} over \DLiteA KBs. It is based on the semantics discussed in \cite{DBLP:conf/rr/LemboLRRS11} where each inconsistency is resolved by removing all data assertions that take part in it and the evaluation was conducted for datasets of few thousand assertions. On the contrary, in \cite{DBLP:conf/bigdataconf/BenbernouO17} the repair is processed on big RDF KBs by only removing one triple from the interdependent inconsistent triples. 

\subsection{Temporal KBs Repair} 
Very few works have investigated the repairing of inconsistent temporal knowledge base. So far, query answering has been extended to the temporal setting in light-weight DLs over inconsistent data, allowing both rigid concepts and roles whose interpretations do not change over time and different types of repair semantics  \cite{DBLP:journals/semweb/BourgauxKT19}. In this work, temporal operators are only used in the definition of queries which are applied to \emph{static} ontologies together with sequences of datasets at time points. The ontology along with the sequence of datasets constitute the temporal knowledge base. Unlike the quoted work, we here use \emph{dynamic} ontologies that allow temporal operators in the definition of concepts. A recent work tackled this issue by translating TDL-Lite KBs into LTL formulas where LTL reasoners can be applied to check their satisfiability \cite{DBLP:conf/dlog/TahratBAGO20}. However, root causes of unsatisfiability, when it exits, are not identified. In this paper, we extend existing approaches of detecting the minimal inconsistent subset in Description Logic (DL) knowledge bases to the temporal setting. We make the assumption that ABox assertions are reliable over time and we focus on repairing the data in the ABox based on the TBox specifications and under the best temporal repair semantic.

%% file: 3-preliminaries.tex
\section{Temporal Description Logic}
\label{sec-tdl}

We now provide details about the syntax and the semantics of the temporal
description logic \TDLLite. In our study as \TDLLite we consider the \TDLLiteN fragment, that allows only future operators interpreted over \Nbb to concepts. We further impose the only use of the two future temporal operators:
$\D_F$(eventually in the future) and $\B_F$(always in the future) and applied only in the right-hand side of inclusions.
%
\begin{definition}{}
Let $\NC, \NI$ and $\NR$ be countable sets of \emph{concept},
\emph{individual names} and \emph{roles} respectively. $\NR$ is the union $\NGl \cup \NL$ where $\NGl$ and $\NL$ are countable and disjoint sets of \emph{global} and \emph{local role names}, respectively. $\TDLLite$ \emph{basic concepts} $B$, \emph{concepts} $C$,\emph{(temporal) concepts} $D$, and \emph{roles} $R$, are formed according to the following grammar:
\[
\begin{array}{llll}
R & ::=  L \mid L^{-} \mid G \mid G^{-}, & B & ::=  \bot \mid \top \mid A \mid, \exists R, \\
C & ::= B \mid \lnot C \mid C_{1} \sqcap C_{2} , & D & ::=C \mid \D_F D \mid \B_F D\mid \lnot D \mid D_{1} \sqcap D_{2}.
\end{array}
\]
where $L \in \NL$, $G \in \NGl$, $A \in \NC$. we called \emph{disjointness}, inclusions of the form $C \sqcap D \sqs\bot$. We also add the ability to specify \emph{functional} roles ($funct~R$).
\end{definition}
%
%
 A \TDLLite \emph{knowledge base}, $\K$, is a pair $\K = (\T,\A)$, where $\T$ is a TBox and $\A$ is an ABox. A \emph{TBox} is a finite set of \emph{general concept inclusions (GCI)} of the form $C \sqs D$ where $C, D$ are \TDLLite concepts, and an \emph{ABox} is a finite set of \emph{concept assertion} of the form $\Next^{n}A(a)$ or $\Next^{n}\neg A(a)$, or a \emph{role assertion} of the form $\Next^{n}R(a,b)$ or $\Next^{n}\neg R(a,b)$, $A\in\NC$, 
$R\in\NR$, $a,b\in\NI$, and\nb{A: added} $n\in \mathbb{N}$.
Here we assume that there is no time point before $0$ or after $n$ and we use abbreviations for assertions $A(a,n)=\Next^n A(a)$ and $R(a,b,n)= \Next^n  R(a,b)$.
\begin{definition}{} \label{def_interpretation}
A \TDLLite \emph{interpretation} is a structure
$\Mmf=(\Delta^{\Mmf}, (\I_{n})_{n \in \mathbb{N}})$, where each
$\I_{n}$ is a classical DL interpretation with non-empty domain
$\Delta^\Mmf$ (or simply $\Delta$).  We have that
$A^{\I_{n}} \subseteq \Delta^\Mmf$ and
$S^{\I_{n}} \subseteq \Delta^\Mmf \times \Delta^\Mmf$, for all
$A \in \NC$ and $S \in \NR$. 
In particular, \textsf{Rigid predicates} are elements from the set of rigid concepts $\textsf{N}_{\textsf{RC}} \subseteq \NC$ or of rigid roles $\NGl$ and for all $X \in \textsf{N}_{\textsf{RC}} \cup \textsf{N}_{\textsf{G}}$ and $i,j \in \Nbb$,
$X^{\I_{i}} = X^{\I_{j}}$ (denoted simply by $X^{\I}$).
Moreover, $a^{\I_{i}}=a^{\I_j}\in \Delta^\Mmf$ for all $a\in\NI$ and
$i,j \in \Nbb$, i.e., constants are \emph{rigid designators} (with
fixed interpretation, denoted simply by $a^{\I}$).
We assume that all interpretations $\Mmf$ satisfy the unique name assumption UNA and the \emph{constant domain assumption} (meaning that objects are not created nor destroyed over time).
%
%
%
The interpretation of roles and concepts at instant $n \in \Nbb$ is
defined as follows (where $S \in \NR$):
\[
\begin{array}{llll}
(S^{-})^{\I_{n}}& = \{ (d', d)\in  \Delta^\Mmf \times \Delta^\Mmf \mid (d, d') \in S^{\I_{n}} \},&
\bot^{\I_{n}} &= \eset, \\
(\exists R.D)^{\Imc_{n}}& = \{d \in \Delta^\Mmf \mid \exists  d' \in D^{\Imc_{n}}: (d,d') \in R^{\Imc_{n}}\},&(\neg D)^{\I_{n}}& = \Delta^\Mmf \setminus D^{\I_{n}},\\
(D_{1} \sqcap D_{2})^{\I_{n}}& = D_{1}^{\I_{n}} \cap D_{2}^{\I_{n}},&(D_{1} \sqcup D_{2})^{\I_{n}} &= D_{1}^{\I_{n}} \cup D_{2}^{\I_{n}}, \\
(\D_F D)^{\I_{n}}& =\bigcup_{k>n} D^{\I_{k}}, & (\B_F D)^{\I_{n}}& =\bigcap_{k>n} D^{\I_{k}},
 \end{array}
\]
\end{definition}
%
\begin{definition}{} \label{def_satisfiability}
We say that a concept $D$ is \emph{satisfied in \Mmf} if there is
$n\in\Nbb$ such that $D^{\I_{n}} \neq \eset$.  The \emph{satisfaction of an axiom in $\Mmf$} is defined as follows:
\[
\begin{array}{llllll}
\Mmf \models D_1\sqsubseteq D_2  &\text{iff}& D_1^{\I_{n}} \subseteq D_2^{\I_{n}} \text{ for all } n\in\Nbb,&&&\\
\Mmf  \models \Next^n A(a)  &\text{iff}& a^{\I} \in A^{\I_{n}}, &
\Mmf \models \Next^n  R(a,b)  &\text{iff}& (a^{\I},b^{\I}) \in R^{\I_{n}}.\\
\Mmf  \models \Next^n \neg A(a)  &\text{iff}& a^{\I} \not\in A^{\I_{n}}, &
\Mmf \models \Next^n \neg R(a,b)  &\text{iff}& (a^{\I},b^{\I}) \not\in R^{\I_{n}}.
 \end{array}
\]
A KB $\Kmc = (\Tmc, \Amc)$ is  \emph{satisfiable} if it exists a model $\Mmf$ that satisfies every axiom of $\Tmc$ and $\Amc$, and written $\Mmf \models \Kmc$.
\end{definition}
\begin{definition}{}
An ABox \Amc ~ is \Tmc-consistent if the KB $\Kmc = (\Tmc, \Amc)$ is \emph{satisfiable}.
\end{definition}

%% file: 4-translation.tex
\section{ Reducing Temporal  \DLite to \DLite }
\label{sec-translation}
This section contains the reduction of $\TDLLite$ KBs, into \DLite KBs. This allows to provide a tight NP-\textit{complete} upper bound for \TDLLite concept satisfiability checking (as shown in section \ref{Trans:m}) and to use highly optimized DL reasoners that can identify the precise set of inconsistent data assertions. The translation is applied on both TBox and ABox levels in sections \ref{trans:TBox} and \ref{trans:ABox} respectively.

\subsection{The Upper Bound }
\label{Trans:m}
The temporal component of our \TDLLiteN fragment, as described in section \ref{sec-tdl}, is based on the propositional LTL, more particularly LTL(F,G) using only the two modal operators $\Rdiamond \equiv F$: sometime in the future and  $\Rbox \equiv G$: always in the future. The main idea is to consider two separate satisfiability problems, one in LTL and the other in \DLite, that together imply satisfiability of \K~in \TDLLiteN.

For the LTL part, it was shown in \cite{DBLP:journals/iandc/DemriS02} that LTL(F,G) is the modal logic S4.3.1 (also called S4.3.Dum or D). Moreover, according to \cite{Nakamura&Ono80}, the satisfiability of a S4.3.1 formula $\phi$ with a maximum nesting depth of modal operators equal to $m$ (maximum temporal depth) is NP-\textit{complete} and there exists a S4.3.1-model which satisfies $\phi$ with at most $m+1$ worlds. Furthermore, there exist an algorithm of a polynomial time complexity for transforming the LTL(F,G)-SAT problem into the SAT problem. As a consequence, a \TDLLiteN satisfiability problem, being NP-\textit{complete}, can be reduced into a \DLite satisfiability problem since $m+1$ is an upper bound of the number of worlds in the model. The reduction is consequently equi-satisfiable to the original \TDLLiteN language. This is confirmed by the case of the upper bound shown by Ladner \cite{Nakamura&Ono80}.

\subsection{TBox Reduction to \DLite }
\label{trans:TBox}
Based on the temporal interpretation $(\I_{n})_{n \in \mathbb{N}}$ of a \TDL KB which is a standard \DL interpretation $\I $ for each time instant (world) $ n\in\Nbb$ as described in definitions \ref{def_interpretation} and \ref{def_satisfiability}, we define the translation in the same way in the interval [0,m]. Given \TDLLite concepts $C$, $D$ and a role $R$, we inductively define the \DLite translation of concepts $C, D$ and a role $R$ at time point $i\in [0,m]$ denoted by tr(C,i,m), tr(D,i,m) and tr(R,i,m) as:
\[
\begin{array}{llll}
 tr(\top,i,m) &=\top,&  tr(\bot,i,m) &=\bot,\\
 tr(C,i,m) &= C_i,&  tr(\neg C,i,m) &= \neg tr(C,i,m),\\ 
 tr(C \sqcap D,i,m)&= tr(C,i,m) \sqcap tr(D,i,m),&  tr(C \sqcup D,i,m)&= tr(C,i,m) \sqcup tr(D,i,m),\\
 tr(\Rbox D,i,m)&= \sqcap_{i}^{m}~tr(D,i,m),&  tr(\Rdiamond D,i,m)&= \sqcup_{i}^{m} tr(D,i,m),  \\
 tr(R,i,m) &= R_{i},& tr(\mathop{\exists R},i,m) &= (\mathop{\exists R_{i}}).
\end{array}
\]
The translation creates fresh concepts $C_i, D_i$ and roles $R_i$ in the $\DLite$ TBox denoting respectively the interpretation $\I_{i}$ of $C, D$ and $R$ at time point $i$. Now, the translation $\T^\dagger$ of a TBox $\T$ is the conjunction of:
 \begin{align}
 \label{eq:tbox}
\bigwedge_{C \sqsubseteq D\in\T} \hspace*{1em}& \bigwedge_{i=0}^{m}\hspace*{0.5em}( tr (C,i,m) \sqsubseteq tr(D,i,m)),\\
 \label{eq:role:global:relation}
\bigwedge_{ R \in\NGl} \hspace*{1em}& \bigwedge_{i=0}^{m} \bigl(tr(R,i,m) \sqsubseteq \sqcap_{i}^{m}~~ tr(R,i,m)\bigr),\\
\bigwedge_{ Funct(R)} \hspace*{1em}& \bigwedge_{i=0}^{m} (Funct(R_i)).
 \end{align}
Note that due the translation of rigid roles (2), the resulting $\T^\dagger$ includes role inclusions. We restrict rigid roles to functional roles in this paper. This extension of \DLite with role inclusions and functional roles is denoted by \DLitebnhn. Its satisfiability problem is \NP~and matches that of the language without role inclusions. However, Local roles are not translated to role inclusions because their semantic is maintained in the resulting \DLite language.

%
\begin{example} \label{ex:2} 
We show the translation of the TBox $\T$ of example \ref{ex:1} where the maximum temporal depth over $\T$ formulas is $m=1$. The resulting \DLite $\T^\dagger$ is:
{\small
\[
\begin{array}{llll}
\T^\dagger=&\{{\sf Adult_0} \sqsubseteq {\sf Person_0},{\sf Adult_1} \sqsubseteq {\sf Person_1},{\sf Minor_0} \sqsubseteq {\sf Person_0},{\sf Minor_1} \sqsubseteq {\sf Person_1}\\
&~~{\sf Person_0} \sqsubseteq {\sf Person_0} \sqcap {\sf Person_1},~{\sf Adult_0} \sqsubseteq {\sf Adult_0} \sqcap{\sf Adult_1},\\
&~~{\sf Minor_0} \sqcap {\sf Adult_0} \sqsubseteq \bot, {\sf Minor_1} \sqcap {\sf Adult_1} \sqsubseteq \bot,\\
&~~{\sf Person_0} \sqsubseteq \exists{\sf hasMother_0},  ~{\sf Person_1} \sqsubseteq \exists{\sf hasMother_1},\\
&~~{\sf Funct(hasMother_0)},~ {\sf Funct(hasMother_1)}\}  
\end{array}
\]}
\end{example} 
\subsection{ ABox Reduction to \DLite }
\label{trans:ABox}
Now, we explain how an ABox $\A$ is translated to $\A^\dagger$. For each $n\in \N$, each concept $A,B \in \NC$ and each role $R, S\in\NR$ , we define:
\begin{equation}
   \label{eq:abox}
  \A^\dagger = \hspace*{-0.5em}
\ \ 
\bigwedge_{\begin{subarray}{c} {\scriptscriptstyle\bigcirc}^n A(a)\in \A \end{subarray}}\hspace*{-1em} A_n(a)\land
\bigwedge_{\begin{subarray}{c} {\scriptscriptstyle\bigcirc}^n R(a,b)\in \A \end{subarray}}\hspace*{-1em} R_n(a,b),
\bigwedge_{\begin{subarray}{c} {\scriptscriptstyle\bigcirc}^n \lnot B(a)\in \A \end{subarray}}\hspace*{-1em} \lnot B_n(a)\land
\bigwedge_{\begin{subarray}{c} {\scriptscriptstyle\bigcirc}^n \lnot S(a,b)\in \A \end{subarray}}\hspace*{-1em} \lnot S_n(a,b),
\end{equation}

$\A^\dagger$ is composed of four conjuncts, the first is the conjunction of the translation of all concept assertions in \Amc and the second is the conjunction of the translation of all role assertions (global and local) occurring in \Amc. The last two conjunctions are equivalent to the first two conjunctions respectively when assertions are negated . 

\begin{example} \label{ex:3} 
We show the translation of the ABox $\A=\A_{1} \cup \A_{2}$ in example \ref{ex:1}. The resulting \DLite $\A^\dagger=\A_{1}^\dagger \cup \A_{2}^\dagger$ is:
{\small
\begin{gather*}
\A_{1}^\dagger=\{{\sf Person_0}({\sf John}), {\sf Minor_1}({\sf John}), {\sf Adult_2}({\sf John}),{\sf Minor_3}({\sf John}),{\sf Minor_4}({\sf John}) \}  \\
\A_{2}^\dagger=\{{\sf hasMother_0}({\sf John},{\sf Ana}), {\sf hasMother_1}({\sf John},{\sf Eva}),{\sf hasMother_2}({\sf John},{\sf Maria}) \}
\end{gather*}
}
\end{example}

It is immediate to verify the satisfiability of the resulting $\Kmc^\dagger=(\T^\dagger,\A^\dagger)$. However, concepts ${\sf Adult_2}$, ${\sf Minor_3}$, ${\sf Minor_4}$ and ${\sf hasMother_2}$ created in $\A^\dagger$ do not occur in $\T^\dagger$ as computed in example \ref{ex:2}. To overcome the above problem, the translated $\T^\dagger$ in the presence of an ABox $\A$ which is defined over a time interval $[l,n]$ should be computed in the interval $[l,n+m]$. 
The translation of $\Kmc=(\T,\A)$ into $\Kmc^\dagger=\T^\dagger\wedge\A^\dagger$ is computed in polynomial time.
\begin{example} \label{ex:4} 
Giving the ABox $\A$ defined over the interval [1,4], the translation of the $\TDLLite$ TBox $\T$ in example \ref{ex:1} into \DLite is computed over the interval [1,5]:
{\small
\[
\begin{array}{llll}
\T^\dagger=&\{
\bigwedge_{i=1}^{5} {\sf Adult_i} \sqsubseteq {\sf Person_i},~~ \bigwedge_{i=1}^{5} {\sf Minor_i} \sqsubseteq {\sf Person_i}, ~~ \bigwedge_{i=1}^{5}{\sf Minor_i} \sqcap {\sf Adult_i} \sqsubseteq \bot,\\ \\
&\bigwedge_{i=1}^{5} {\sf Person_i} \sqsubseteq (\sqcap_{i}^{5} {\sf Person_{i}}),~ \bigwedge_{i=1}^{5} {\sf Adult_i} \sqsubseteq (\sqcap_{i}^{5} {\sf Adult_{i}}),\\ \\
& \bigwedge_{i=1}^{5} {\sf Person_i} \sqsubseteq \exists{\sf hasMother_i},
~~\bigwedge_{i=1}^{5} ({\sf hasMother_i} \sqsubseteq \sqcap_{i}^{5} {\sf hasMother_{i}}), \\ \\
&\bigwedge_{i=1}^{5} {\sf Funct(hasMother_i)}\} 
\end{array}
\]}
\end{example} 

\begin{theorem}\label{lem:qtli-equisat}
A \TDLLiteN KB $\Kmc$ is satisfiable iff the \DLitebnhn-formula $\Kmc^\dagger$ is satisfiable. Moreover, $\Kmc^\dagger$ can be constructed in polynomial time w.r.t. the size of \Kmc.
\end{theorem} \nb{TODO: the same Proof}
\begin{proof}
Theorem~\ref{lem:qtli-equisat} is proved in the same way as the standard translation to FOL plus Theorem 6 in \cite{Nakamura&Ono80} .
\end{proof}
%

%
%
%
%
%
%
%
%


%% file: 5-detection.tex
\section{Inconsistency Detection and Repair in TKBs}
Once $\TDLLite$ KBs are mapped into $\DL$ KBs as defined in the previous section, the following step is to detect the inconsistent data assertions using $\DL$ reasoners. In this section, we first recall the essentials of the DL inconsistency detection and then present the core procedure of computing the minimal inconsistent subset and the best repair in \TDLLite.  


\subsection{Inconsistency Detection in KBs}

The Web Ontology Language OWL is a logic-based language of knowledge representation intended to be used to verify the consistency of a dataset with the semantics of the underlying knowledge representation formalism. This task is usually handled by what we call OWL reasoners such as RacerPro\footnote{\url{https://github.com/ha-mo-we/Racer}}, Pellet\footnote{\url{http://pellet.owldl.com/}}, FaCT++\footnote{\url{http://owl.cs.manchester.ac.uk/tools/fact/}} and others. In FaCT++, nominals are unavailable and ABox reasoning is not supported. Nominals are also unvailable in RacerPro. Therefore, we use Pellet as our OWL reasoner: it is the first sound and complete tableau-based reasoner for the OWL-DL sublanguage (a syntactic variant of the Description Logic \Shoin \cite{DBLP:journals/ws/SirinPGKK07} which is much more expressive than the tractable \DLitebnhn). Pellet is written in Java, open source and efficient when the number of  instances is large. Moreover, It also offers a specific service for computing inconsistency explanations on the TBox terminology and the assertional ABox levels. Moreover, Pellet outperforms RacerPro when reasoning  on a large number of instances  \cite{DBLP:journals/ws/SirinPGKK07}.

Before describing the inconsistency detection approach, we should explain what "inconsistent data assertion" means in the context of \DLitebnhn. We can distinguish three different types of \DLitebnhn TBox constraints: disjointness which are GCIs of the form ($C \sqcap D \sqsubseteq \bot$), functionality assertions of the form (funct R) and rigid predicates. 


\begin{figure*}[t]
  \centering
  \includegraphics[width=0.95\linewidth]{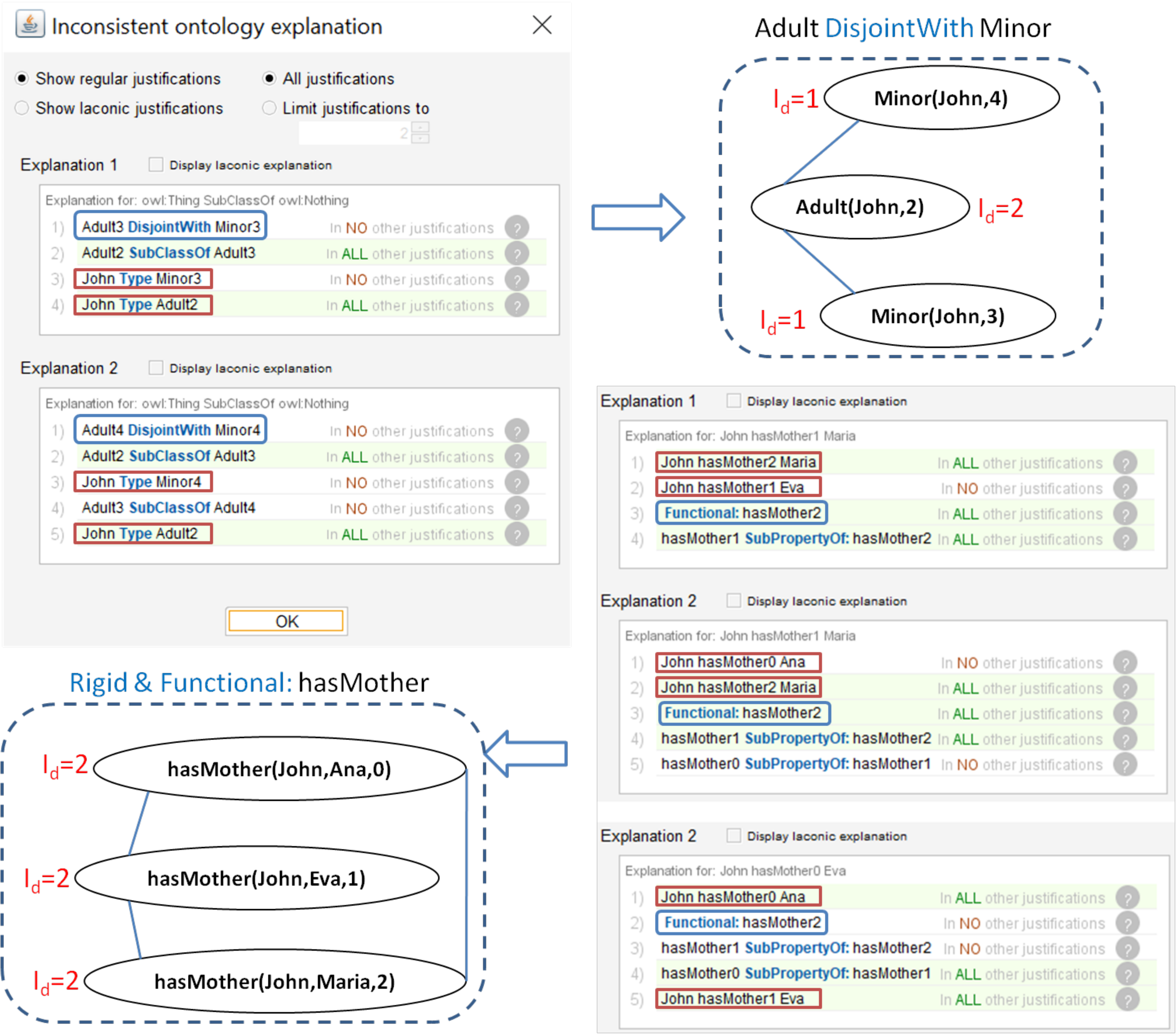}
  \caption{Inconsistency detection and explanation over example \ref{ex:1} giving by Pellet via the Protegé \footnote{https://protege.stanford.edu/} Ontology editor. We give the corresponding Inconsistency Graph to show conflicts between the inconsistent assertions. We report the number of conflicts (number of edges) of each inconsistent assertion in the Inconsistency degree $I_d$.}
  \label{fig:Pellet Expl}
 \end{figure*}


\begin{definition}
Let $\Kmc^\dagger=(\T^\dagger,\A^\dagger)$ be a \DLitebnhn KB and let c be a disjointness inclusion or a functionality assertion or a rigid predicate of $\T^\dagger$.~A set of data assertions $I=\langle a_1, a_2,... \rangle \in \A^\dagger$, is called inconsistent, iff there is some $c \in$ $\T^\dagger$,~such that $I \nvDash c$.
\end{definition}

Our approach starts by checking the satisfiability of the translated TBox $\T^\dagger$ then checks the consistency of the translated ABox $\A^\dagger$ according to $\T^\dagger$ using Pellet. If $\A^\dagger$ is inconsistent, the explanation support of Pellet points the inconsistent set of data assertions without a resolution strategy. However, this explanation support is an axiom tracing service which allows to extract, from the TBox and the ABox, the relevant axioms responsible for the inconsistency which can directly be used for reparation.

Based on the resulting \DLitebnhn KB $\Kmc^\dagger$ in section \ref{trans:ABox} of the motivating example, all the inconsistency explanations provided by Pellet are shown in figure \ref{fig:Pellet Expl}. First, we can observe inconsistency due to conflicts in axioms: {\small(${\sf John~ Type~ Adult_2}$)} with {\small(${\sf John~ Type~ Minor_3}$)} and again {\small(${\sf John~ Type~Adult_2}$)} with {\small(${\sf John~ Type ~Minor_4}$)} according to the disjointness CIs {\small(${\sf Adult} \sqcap {\sf Minor} \sqsubseteq \bot$)} translated to {\small(${\sf Adult_3~ DisjointWith ~Minor_3}$)}  and to {\small(${\sf Adult_4~ DisjointWith~ Minor_4}$)} respectively at time point $3$ in explanation $1$ and at time point $4$ in explanation $2$. We also notice that axioms can be responsible for multiple conflicts according to the constraints in $\T^\dagger$. The axiom {\small$({\sf John Type~Adult_2}$)} which corresponds to the assertion {\small{\sf Adult(John,2)}} in \A~ has two conflicts according to explanations 1 and 2 in the left of figure \ref{fig:Pellet Expl}. Formally, we define an \emph{Inconsistency degree} $I_d$ as the number of $\T^\dagger$ constraints (conflicts) on which an assertion is involved.
Second, considering the rigid role {\small{\sf HasMother}}, we can observe from the Pellet explanation, on the right-side of figure \ref{fig:Pellet Expl}, inconsistencies due to conflicts between axioms: {\small(${\sf John ~hasMother_0~Ana}$)}, {\small(${\sf John~hasMother_1~Eva}$)} and {\small(${\sf John~hasMother_2~Maria}$)} according to the functional and rigid role {\sf hasMother}. Consequently, each of these assertions has an $I_d=2$. To report these inconsistent assertions in an intuitive way, we use an \emph{ Inconsistency graph} which is similar to the \emph{conflict-hypergraph} used to represent constraint violations in databases \cite{DBLP:journals/iandc/ChomickiM05} or in inconsistency DL setting \cite{DBLP:conf/kr/BienvenuB20,DBLP:conf/bigdataconf/BenbernouO17}.

\begin{definition}
Let $\Kmc^\dagger=(\T^\dagger,\A^\dagger)$ be a \DLitebnhn KB and let $I_m=\langle a_i, a_j,... \rangle \nvDash c_k$ be sets of inconsistent data assertion where $c_k \in \T^\dagger$-constraints. An Inconsistency Graph of $\Kmc^\dagger$ is an edge-labeled graph denoted by $IG(K)=(V,E)$ such that for all $I_m$, $V=\{a_i|a_i \in I_m\}$ and $E=\{ (a_i, a_j)| \langle a_i, a_j\rangle \nvDash c_k, c_k \in \T^\dagger\}$
\end{definition}

This graph is built by iterating over all Pellet explanations $I_m$ as follows:
\begin{itemize}
  \item for every assertion $a_i$ of the form ({\sf a Type $D_i$}) or ({\sf a $R_i$ b}), we add the vertice ({\sf D(a,i)}) or ({\sf R(a,b,i)});
  \item for every pair of vertices in $I_m$ , we add an edge connecting them which we label with the broken constraint.
\end{itemize}
The inconsistency graph of the motivated example is shown in figure \ref{fig:Pellet Expl}. Note that we can easily determine how many conflicts in which each inconsistent assertion is involved. This graph is used as an input for the repair phase which provides graph theories and tools useful for repair purposes.
\input{6-reparation.tex}

%% file: 6-reparation.tex
\subsection{Best Repair in KBs}
\label{temporal repair}
After building the inconsistency graph which encodes the inconsistent data assertions, the next step is to repair them. An extreme solution would be to simply throw away all the detected inconsistent assertions from \Amc. This would certainly not meet the expected repair requirement which consists of applying a minimal set of changes that restore consistency. Typically, minimality is defined by a set of inclusion yielding:

\begin{definition}
Consider a \TDLLiteN KB $\Kmc = (\Tmc, \Amc)$. A temporal ABox repair of \Tmc is a set $\Amc^\prime$ of assertions such that:~(i) $\Amc^\prime \subseteq \Amc$; (ii) $\Kmc = (\Tmc, \Amc^\prime)$ is consistent; (iii) $ \A' \subset \A'' \subseteq \A$ where $\Kmc = (\Tmc, \A'')$ is consistent and $\A''$ does not exist.
\end{definition}

In other words, $\A'$ is a maximal consistent subset of $\A$ that is obtained by throwing away a minimal set of inconsistent assertions (aka the Minimal Unsatisfiable Set MUS). This is performed by the removal of one of the two inconsistent data assertions involved in the conflict. In the inconsistency graph, this corresponds to removing one of the two vertices that are connected by the edge representing this conflict. A complete repair is in fact the well-known problem of finding the minimum vertex cover \cite{DBLP:journals/talg/Karakostas09} which computes a set of vertices (a MUS) whose removal leads to the removal of all the edges (all the conflicts) of the inconsistency graph.

Recall that the computation of the minimum vertex cover is a classical NP-\textit{complete} problem. However, an approximation algorithm, such as the 2-approximation algorithm in \cite{DBLP:journals/talg/Karakostas09} can be applied in a greedy manner until there are no more edges in each connected component of the inconsistency graph as follows:
\begin{itemize}
  \item for each step we select the vertex cover which is the vertex having a higher inconsistency degree $I_d$ (the most inconsistent assertions are those involved in most conflicts) ;
  \item If more than one vertice have the same degree, one is randomly selected as vertex cover.
\end{itemize}
The union of vertex cover sets of the connected components of the graph forms a vertex cover (a MUS) of the entire inconsistency graph.

For instance, let's consider the inconsistency graph of figure \ref{fig:Pellet Expl}. The minimum vertex cover algorithm will compute the repairs in the first connected component of the graph labeled by the constraint {\small(${\sf Adult~ DisjointWith ~Minor}$)} by removing the assertion {\small{\sf Adult(John,2)}} because it has the highest $I_d$. In the second connected component labeled by the constraint {\small(${\sf Rigid\&Fuctional:~hasMother}$)}, the repair will be computed by removing randomly two of the three assertions in the graph as they have the same $I_d$.

Let us note that there might be several possible MUSs for the same inconsistency graph, since some vertices can have the same degree of inconsistency. Therefore, by randomly removing one of the two vertices, we obtain a different MUS but all possible MUSs are minimal. The repair is then always maximal and is obtained by removing the resulting MUS from the ABox. 


\begin{figure*}[t]
\begin{subfigure}{.5\textwidth}  
  \centering
  \includegraphics[width=.8\linewidth]{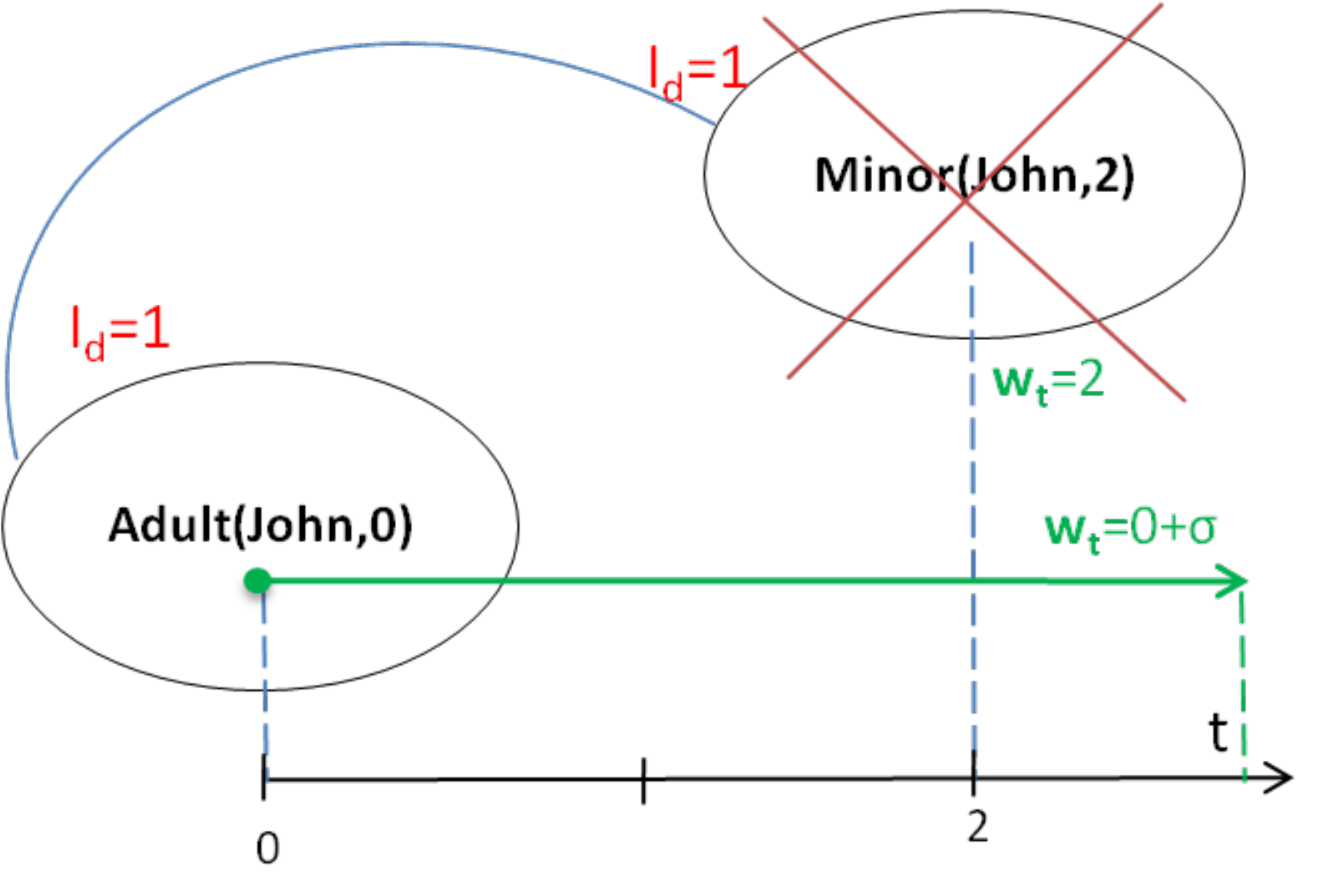}  
  \caption{case where the recent assertion is removed}
  \label{fig:sub-a}
\end{subfigure}
\begin{subfigure}{.5\textwidth}
  \centering
  \includegraphics[width=.8\linewidth]{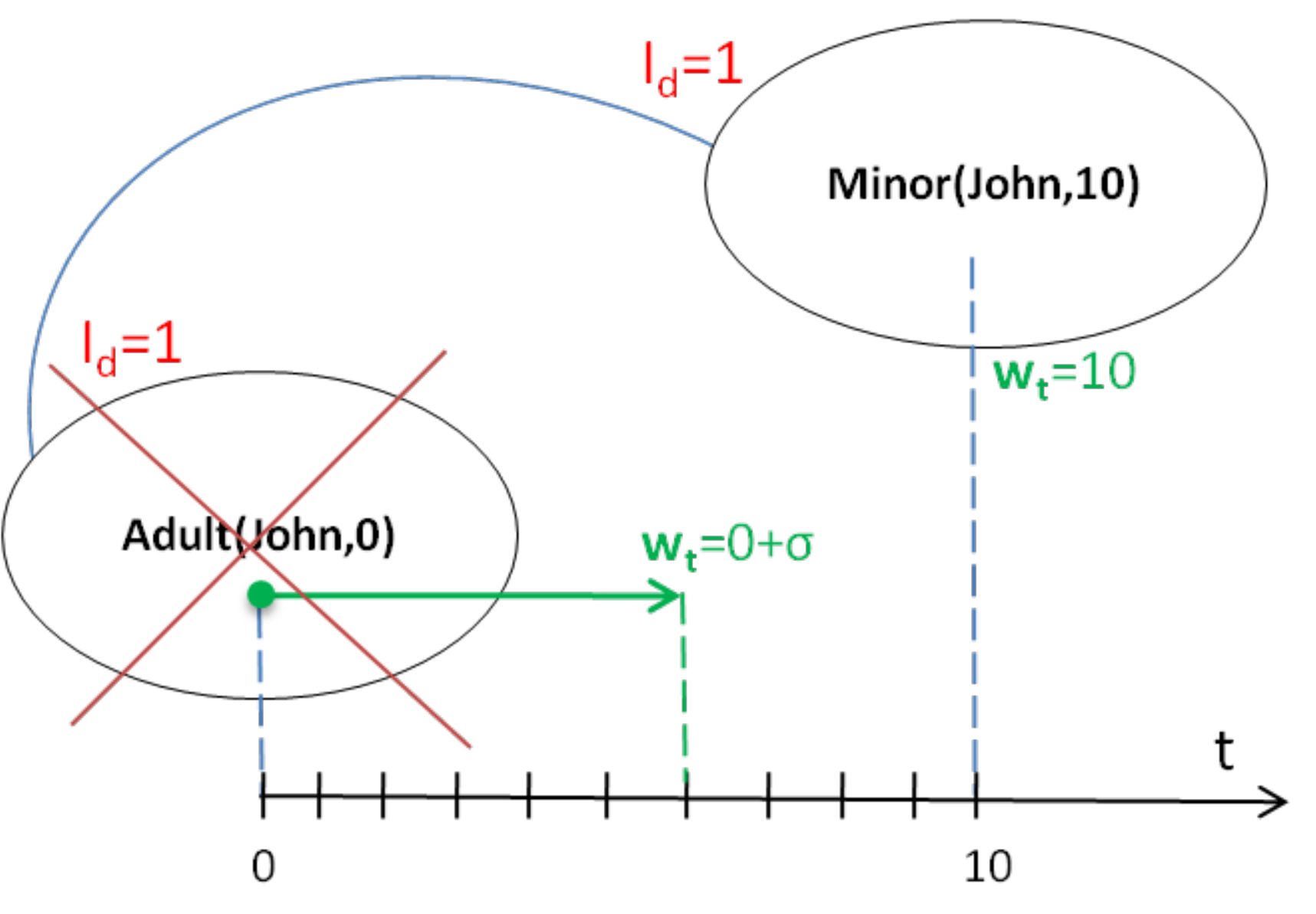}  
  \caption{case where the rigid assertion is removed}
  \label{fig:sub-b}
\end{subfigure}
\caption{Best temporal repair in TKBs}
\label{fig:wt}
\end{figure*}

\subsection{Best Temporal Repair in TKBs}
\label{sigma}
Clearly, not just any repair is useful or interesting in the temporal setting. For instance, repairs that return only ancient assertions might be unwanted. However, in the light of temporal knowledge bases, we are interested in  defining a strong notion of temporal repair. For such a purpose, we associate for each inconsistent assertion a temporal weight by taking the defining rigid axioms and the freshness of the assertion into consideration. Our aim is to guide the repair algorithm to remove assertions with the lowest temporal weight when they have the same degree of inconsistency.
\begin{definition}
We assign a temporal weight $w_t=t_i+\sigma$ for each inconsistent assertion $a_i$ associated with the timestamp $t_i$. If $a_i$ is an instance of a rigid predicate, $\sigma$ expresses a duration 
for this rigid predicate. Otherwise, the temporal weight of $a_i$ is based only on the timestamp $t_i$ ($\sigma=0$), so expresses the freshness of the assertion, as follow:
$$ w_t(a_i, t_i) = \left\{
    \begin{array}{ll}
        t_i+ \sigma & \mbox{if } a_i \in \textsf{N}_{\textsf{RC}} \cup \textsf{N}_{\textsf{G}}  \\
        t_i & otherwise
    \end{array}
\right.
$$
\end{definition}
The intuition behind using a time range $\sigma$ for each rigid predicate in repair phase is to set a maximum time threshold after which the assertion of this rigid predicate is discriminated or weakened. 
Figure \ref{fig:wt} shows that the classic maximal repair in cases \ref{fig:sub-a} and \ref{fig:sub-b} could be the same because they share the same $I_d$. However, it is easy to see on a timeline that it is better to remove {\small${\sf Minor(John,2)}$} in (\ref{fig:sub-a}) and {\small${\sf Adult(John,0)}$}  in (\ref{fig:sub-b}): the notion of temporal weight is intended to capture situations where a maximal repair is temporally better than an other. At this level, we are considering a semi-automatic approach, which is guided by a user who will fix the value of $\sigma$ by giving a duration to each defined rigid predicate according to his preferences, thus providing a repair that is as close as possible to the user needs.

%% file: 7-conclusion.tex
\section{Conclusion and Future Work}
\label{sec:conclusion}

This paper provides a first exploration of repairing the ABox w.r.t a TBox defined over Temporal \DLite. The temporal language considered so far is the \TDLLiteN with which we can express and check several useful types of temporal constraints, such as defining temporal concepts in GCIs and rigid predicates, while maintaining good computational features.
The first step in the reparation process is the detection of inconsistencies in \TDLLite KBs. To do so, we proposed an equisatisfiable translation from \TDLLite into \DLite KBs in order to use well optimized DL reasoners that include an axiom tracing service which allows extracting, from the TBox and the ABox, the relevant axioms involved in the inconsistency. This allows, in the second step, to perform a reparation based on the best repair semantic over \DLite ABoxes. We extended this semantic to the temporal setting by defining a temporal weight to guide the repair by removing, from the ABox, assertions with the highest inconsistency degree and the lowest temporal weight. Repair computation can be performed in polynomial time with respect to the number of inconsistent data assertions that appear in the ABox.

As a direction for our future work, we aim to enrich the definition of the TBox with General Concept Inclusions GCIs having temporal past operators on the right-hand side of the GCI. This could be equivalent to having future temporal operators on the left hand side of the GCI like $\{D \sqsubseteq C \text{, where D is a temporal concept}\}$. Let us note that in LTL some future temporal operators when expressed in the left hand side of an inclusion such as $(\D_F~q \to  p) $ can be expressed as $(q \to \B_P~p)$ using a past operator on the right-hand side of the inclusion. More generally, we plan to investigate in practice repairing KBs based on multiple combinations of LTL with DL-Lite logics which are First Order rewritable \cite{DBLP:conf/dlog/ArtaleKKRWZ14}.
Also, in the same spirit of the proposed temporal weight $\sigma$ in section \ref{sigma}, which we defined as a temporal range for rigid predicates, we are considering adding metric operators to the \TDLLite language \cite{DBLP:conf/ecai/Gutierrez-Basulto16, DBLP:journals/tocl/BaaderBKOT20} that augment LTL temporal operators with time interval. Finally, it would be interesting to implement a repair framework and evaluate the scalability properties of our approach based on the temporal best repair semantics against temporal query answering under other semantics.